\documentclass{article}
\usepackage{spconf,amsmath,graphicx}
\usepackage{multirow} 
\usepackage{amssymb}
\usepackage{url}

\title{SIMI: Self-information Mining Network for Low-light Image Enhancement}
\name{Xuanshuo Fu, Lei Kang\dag, Javier Vazquez-Corral \thanks{This work has been partially supported by the predoctoral program AGAUR-FI ajuts (2026 FI-3 00470) Joan Oró, which are backed by the Secretariat of Universities and Research of the Department of Research and Universities of the Generalitat of Catalonia, as well as the European Social Plus Fund; the Beatriu de Pinós del Departament de Recerca i Universitats de la Generalitat de Catalunya (2022 BP 00256), and the Grant PID2024-162555OB-I00 funded by MCIN/AEI/10.13039/501100011033, ERDF ``A way of making Europe''. JVC also acknowledges the 2025 Leonardo Grant for Scientific Research and Cultural Creation from the BBVA Foundation. The BBVA Foundation accepts no responsibility for the opinions, statements and contents included in the project and/or the results thereof, which are entirely the responsibility of the authors. \\ \dag denotes Corresponding Author.}}
\address{Computer Vision Center, Barcelona, Spain \\  Computer Science Dept., Universitat Autònoma de Barcelona, Barcelona, Spain\\
\tt \small \{xuanshuo, lkang, javier.vazquez\}@cvc.uab.cat}
\begin{document}
%
\maketitle
\begin{abstract}
Poor lighting conditions significantly impact image quality, posing substantial challenges for image editing and visualization. Many existing enhancement methods aim at proposing complex models while neglecting the intrinsic information contained within low-light images. In this work, we propose the Self-Information Mining (SIMI) network, an innovative unsupervised framework that decomposes low-light images into multiple components based on bit-plane decomposition. Our approach allows mining intrinsic information without relying on external data. This not only accelerates model convergence but also improves performance and reduces computational overhead. The unsupervised nature of our method facilitates real-world applicability.  Experiments conducted on standard benchmarks demonstrate that SIMI achieves state-of-the-art performance. \url{https://github.com/casted/SIMI}.
\end{abstract}

\section{Introduction}
\label{sec:intro}

Low-light image enhancement (LLIE) aims to improve the perceptual and functional quality of images captured under insufficient illumination. Such images suffer from low brightness, poor contrast, amplified noise, and color distortion, degrading both visual quality and performance in downstream tasks, such as object detection and classification.

Early methods employed handcrafted priors such as histogram equalization, gamma correction, and Retinex theory \cite{jain1989fundamentals, land1971lightness, jobson1997multiscale}. While interpretable, they often fail in complex lighting, producing over-enhancement, halos, or unnatural colors.

Deep learning has since transformed LLIE. Supervised models \cite{wei2018deep, zhang2019kindling, cai2023retinexformer, jiang2021enlightengan} use CNNs, GANs, or Transformers to learn direct mappings. Recent advances include SNR-aware Transformers \cite{xu2022snr}, structure-guided frameworks \cite{xu2023low}, and unified Retinex architectures \cite{cai2023retinexformer}. Generative approaches like normalizing flows \cite{wang2022low} and diffusion models \cite{wang2023exposurediffusion} model conditional distributions or exposure priors. Yet these methods require large paired datasets, are computationally heavy, and generalize poorly.

To mitigate these issues, unsupervised (also called zero-shot) methods have emerged. Zero-DCE and Zero-DCE++ \cite{guo2020zero, li2021learning} estimate enhancement curves via no-reference metrics. RUAS \cite{liu2021retinex} integrates Retinex with self-supervised learning, while SCI \cite{ma2022toward} uses cascaded self-calibration. PairLIE \cite{fu2023learning} enforces reflectance consistency from paired captures, and EnlightenGAN \cite{jiang2021enlightengan} leverages unpaired adversarial learning. Diffusion-based methods like GDP \cite{fei2023generative} and FourierDiff \cite{lv2024fourier} exploit generative priors but risk hallucinations or require fidelity constraints.

\begin{figure*}[ht!]
  \centering
 \vspace{-8mm}
  \includegraphics[width=0.98\linewidth]{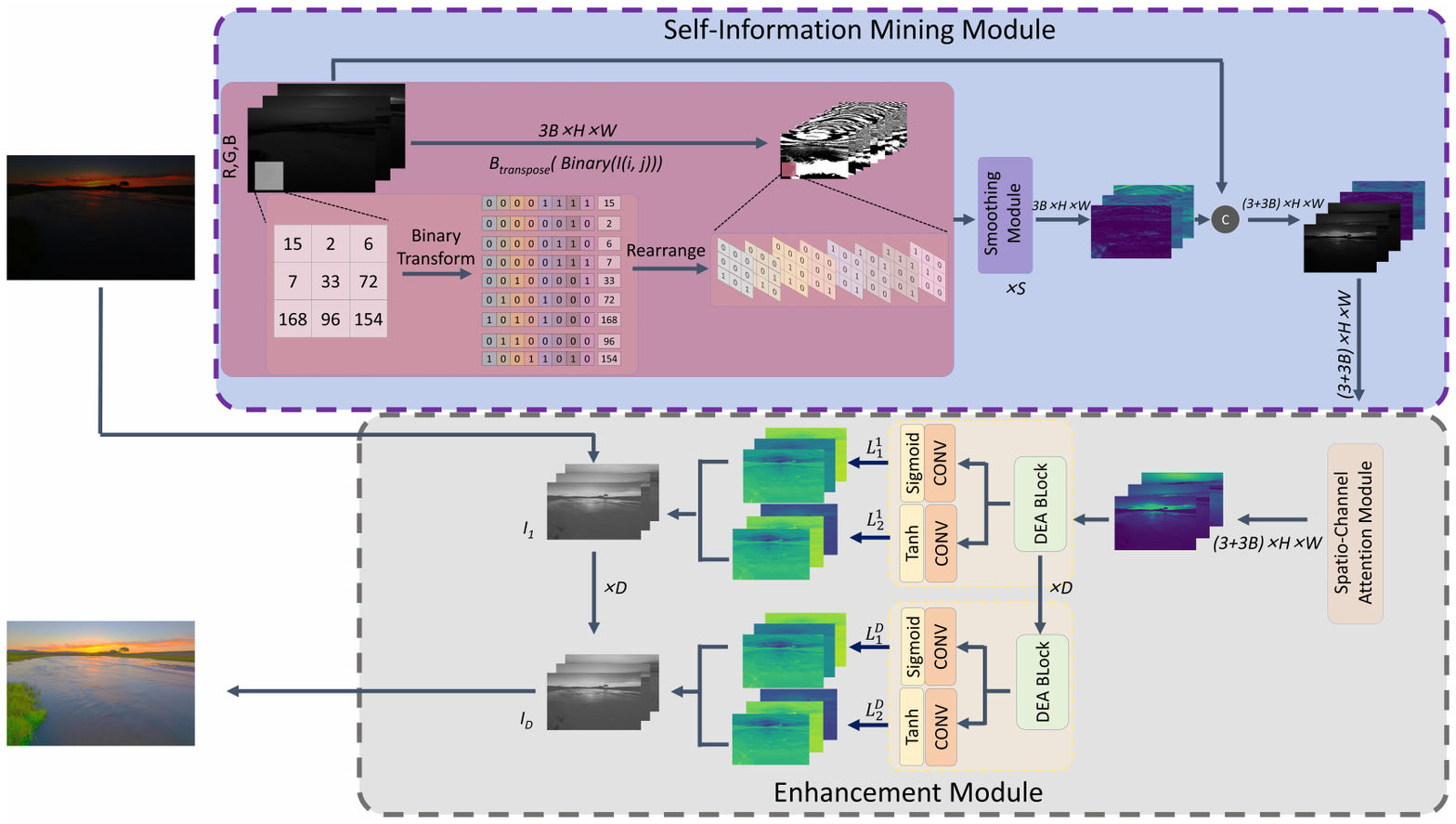}
   \vspace{-18mm}
  \caption{The proposed architecture comprises two main components: a self‑information mining module (blue) and an enhancement module (grey). Within the self‑information module, the input image is first decomposed into bit‑plane channels (red block). These channels are then processed by a set of smoothing units and concatenated with the original image. The enhancement module contains a spatio‑channel attention block followed by a sequence of DEA blocks, which extract illumination and structural cues. These cues are finally applied to the input RGB image to generate the enhanced output.}
     \vspace{-4mm}
  \label{fig:net}
\end{figure*}

To tackle generalization, efficiency, and structural fidelity in unsupervised LLIE, we propose the \textbf{Self-Information Mining (SIMI)} network. SIMI extracts intrinsic cues via bit-plane decomposition and fuses them with the input in a lightweight architecture for enabling adaptive, content-aware enhancement without paired data. Our contributions are:

\begin{enumerate}
    \item A novel self-information mining mechanism via bit-plane decomposition, uncovering latent enhancement cues without supervision or pretraining.
    \item A lightweight, end-to-end unsupervised model that fuses mined cues with input for accurate, adaptive enhancement at low computational cost.
    \item Experiments on three different datasets demonstrate that the proposed method obtains state-of-the-art results, showing strong generalization across diverse real-world low-light conditions.
\end{enumerate}

\section{Methodology}

Figure \ref{fig:net} presents the proposed network architecture. The low‑light input is first decomposed into bit‑plane maps (red box), which reveal self‑information such as boundaries and texture details typically hidden under low‑illumination conditions. A spatio-channel attention mechanism then allocates adaptive channel and pixel‑level weights to balance the contribution of each layer. Finally, the cascaded DEAblocks extract illumination and structural cues that guide the enhancement of the original RGB image. Additional details are provided below.

\subsection{Self-Information Mining Module}

\begin{figure}[t!]
  \centering
  \includegraphics[width=\linewidth]{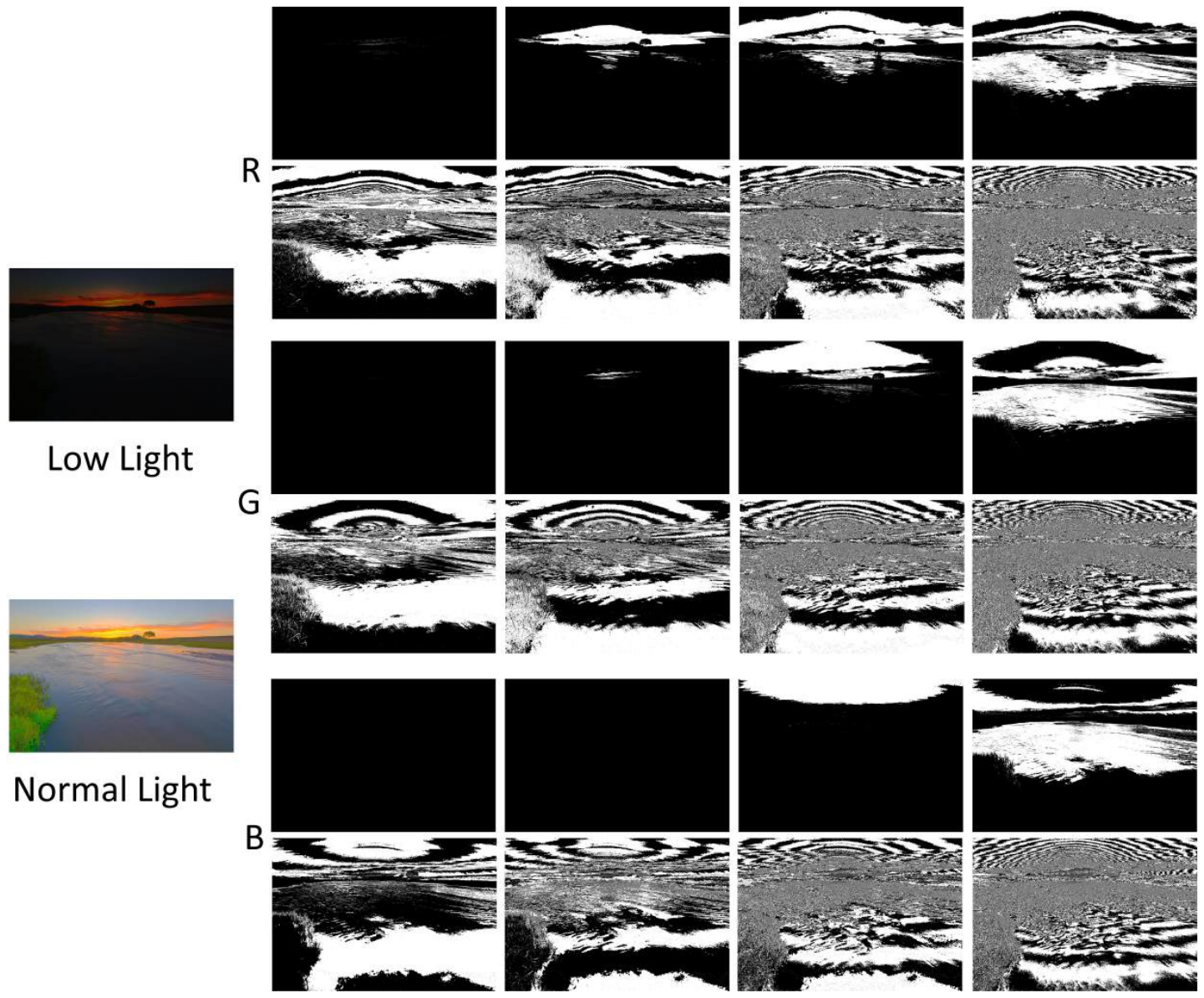}
   \vspace{-8mm}
  \caption{Example of a bit‑plane decomposition of an image. The original RGB image is decomposed into bit‑plane maps for each channel, allowing us to independently expose information that is otherwise hidden at lower significance levels. This representation reveals subtle structures and details not visible in the original image, providing the network with a richer and more informative input for the enhancement.}
  \label{fig:RGB Decomposition}
\end{figure}

We begin by extracting self‑information through a bit‑plane decomposition of the $H\times W$ RGB input. Each channel of the 8‑bit image is converted into its binary representation, producing $B=8$ bit‑plane maps per channel and, thus, resulting in a tensor of size $H\times W\times 3B$ (Fig.~\ref{fig:RGB Decomposition}). This representation exposes latent cues—such as boundaries, textures, and subtle intensity variations—that are often difficult to perceive in low‑light images. The bit‑plane representation also introduces a natural hierarchy: higher planes encode coarse structural information, while lower planes capture fine details and noise. Exploiting this hierarchy enables effective multi‑level feature learning for low‑light enhancement.

To mitigate discretization artifacts and refine the extracted signals, we apply $S$ smoothing modules, each composed of a convolutional layer followed by a SiLU activation. The refined bit‑plane features are concatenated with the original RGB image and then forwarded to the enhancement module.

\subsection{Enhancement module}
First, the features are processed by a CBAM‑style spatio–channel attention module~\cite{woo2018cbam}, which adaptively reweights channel and spatial responses to emphasize informative components while suppressing irrelevant ones.

Next, the attended features are processed by a sequence of $D$ cascaded DEABlocks~\cite{chen2024dea}. At stage $i$, each block predicts an illumination‑related feature $\mathbf{L}_1^{i}$ and a complementary structural feature $\mathbf{L}_2^{i}$. These predictions are then used to recursively update the intermediate enhanced image:
\begin{equation}
\resizebox{.9\hsize}{!}{$\displaystyle
\mathbf{I}_i = \mathbf{I}_{i-1} + \mathbf{I}_{i-1} \cdot (\mathbf{L}_1^{i-1} - \mathbf{I}_{i-1}) \cdot
\underbrace{\frac{\mathbf{L}_1^{i-1}}{\sigma(-\mathbf{I}_{i-1} + \mathbf{L}_2^{i-1} - 0.1)\cdot \mathbf{L}_2^{i-1}}}_{\text{adaptive modulation}},\quad
\sigma(x)=\frac{1}{1+e^{-10x}}
$}
\end{equation}
where $\mathbf{I}_0$ denotes the input image and $\mathbf{I}_D$ represents the final enhanced output. The sigmoid gate $\sigma(\cdot)$ modulates the enhancement strength based on local exposure, allowing the model to adaptively adjust different brightness regions. Meanwhile, the interaction with $\mathbf{L}_2^{i-1}$ onstrains the update to preserve structural details and prevent over‑enhancement artifacts. Overall, this recursive formulation balances illumination correction with structural fidelity, in line with zero‑reference enhancement approaches~\cite{guo2020zero}. 

\begin{figure*}[t!]
\vspace{-3cm}
  \centering
  \includegraphics[width=\linewidth]{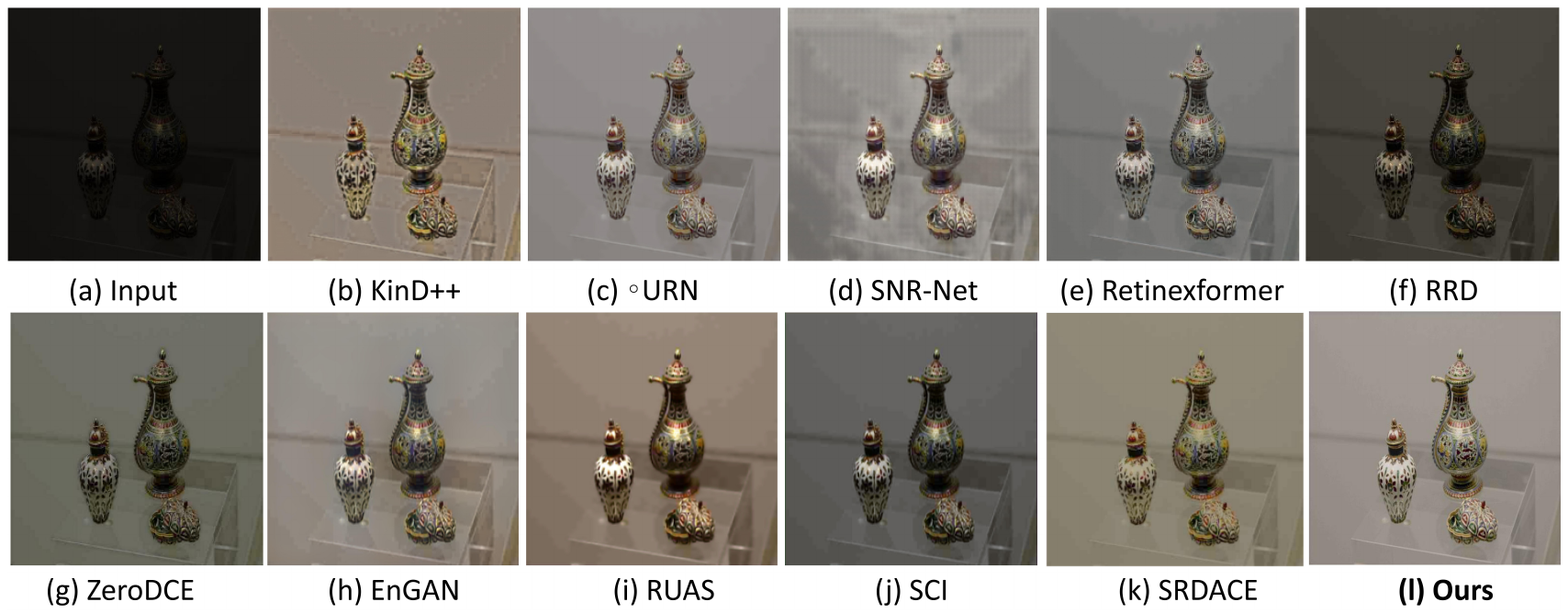}
   \vspace{-3.5cm}
  \caption{Comparison of qualitative results across methods. Our method generates smoother reconstructions with fewer noise artifacts, as evidenced by the more stable color appearance of the vase.}
   \vspace{-4mm}
  \label{fig:CMP}
\end{figure*}

\subsection{Loss}


We primarily adopt the unsupervised loss functions proposed in \cite{wen2023self}, which consist of four main components.
\vspace{3mm}

\noindent \textbf{Localized Color consistency.} This loss enforces consistency between the color distributions of the input and output images and is defined as
\begin{equation}
    \mathcal{L}_{lc} = \sum_{c\in{RGB}} ||F^c_{in} - F^c_{ou}||,
\end{equation}
where $F^c_{in}$ and $F^c_{ou}$ denote the pixel-wise color estimation factors of the input and output images, respectively. The factor map is obtained by normalizing each color channel of the input image by the sum of all channels, with a small offset added to avoid division by zero.

\vspace{3mm}
\noindent \textbf{Global Chromatic Fidelity loss.} To prevent color over-saturation, we employ the global chromatic fidelity loss:
\begin{equation}
    \mathcal{L}_{g} = \sum_{c\in{RGB}} \left(\varphi^c_{ou} - \frac{1}{3}\right)^2,
\end{equation}
where $\varphi^c_{ou}$ denotes the spatial average of channel $c$ in the output image. This loss is motivated by the gray-world assumption and penalizes over-saturated results.
\vspace{3mm}

\noindent \textbf{Brightness Preservation loss.}
This loss aims at maintaining overall luminance consistency. It is mathematically formulate as 
\begin{equation}
    \mathcal{L}_{lu} = ||\mathcal{B} - \sum_{c\in{RGB}} I^c_{ou}||^2_2,
\end{equation}
where $\mathcal{B}$ represents the expected brightness of the output image. Specifically, $\mathcal{B}$ is computed by multiplying the ground-truth by $3$ and scaling it by one minus the sum of the $L_2$ norms of the differences between $F^c_{in}$ and $\frac{1}{3}$ across the three color channels. Here, $I_{ou}$ denotes the output image produced by our method.

\noindent \textbf{Global Chromatic Fidelity loss.}
Finally, we include a curve regularization term to encourage smoothness:
\begin{equation}
    \mathcal{L}_{s_j} = \frac{1}{N}\sum_{c\in{RGB}} ||\nabla \vartheta_j^c||^2_2,
\end{equation}
where $N$ is the number of pixels and $\vartheta_1 \in \{L_1\}$, and $\vartheta_2 \in 
\{L_2\}$ denote the two curve representations used in our model.

\noindent \textbf{Total loss.} The overall training objective is a weighted combination of the above losses:
\begin{equation}
    \mathcal{L}_{total} = \alpha \mathcal{L}_{lc} + \beta \mathcal{L}_{g} + \gamma \mathcal{L}_{lu} + \delta_1 \mathcal{L}_{s_1} + \delta_2 \mathcal{L}_{s_2},
\end{equation}
where $\alpha$, $\beta$, $\gamma$, $\delta_1$ and $\delta_2$ are weight adjustment factors.

\section{Experiments}
\subsection{Dataset}

\begin{table*}[t]
\centering\caption{Performance comparison with state-of-the-art methods. All models are trained on SCIE Part I and evaluated directly on LOLV1, LSRW, and SCIE Part II. Reported results are averaged over all test images.}
\vspace{0.1cm}
\resizebox{\textwidth}{!}{
\begin{tabular}{cccccccccccc}
\hline
 & & \multicolumn{3}{c}{\textbf{Supervised Learning}} &  &\multicolumn{6}{c}{\textbf{Unsupervised Learning}}\\\cline{3-5} \cline{7-12}
\textbf{Dataset} & \textbf{Metric} & \textbf{Retinex-Net~\cite{wei2018deep}} & \textbf{KinD~\cite{zhang2019kindling}} & \textbf{KinD++~\cite{zhang2021beyond}} &  &\textbf{EnlightenGan~\cite{jiang2021enlightengan}} & \textbf{ZeroDCE~\cite{li2021learning}} & \textbf{RUAS~\cite{liu2021retinex}} & \textbf{SCI~\cite{ma2022toward}} & \textbf{SRDACE~\cite{wen2023self}} & \textbf{Ours} \\
\hline
\multirow{3}{*}{LOLV1} & PSNR$\uparrow$ & 16.77 & 17.65 & 17.75 & & 17.48 & 14.86 & 16.4 & 14.78 & 18.50 & \textbf{18.89} \\
 & SSIM$\uparrow$ & 0.42 & \textbf{0.77} & \textbf{0.77} & & 0.65 & 0.56 & 0.50 & 0.52 & 0.58 & 0.60 \\
 & LPIPS$\downarrow$ & 0.47 & \textbf{0.18} & 0.20 & & 0.32 & 0.34 & 0.27 & 0.34 & 0.33 & 0.25 \\
\hline
\multirow{3}{*}{LSRW} & PSNR$\uparrow$ & 15.51 & 16.41 & 16.08 & & 17.08 & 15.86 & 14.27 & 15.24 & 17.10 & \textbf{17.35} \\
 & SSIM$\uparrow$ & 0.37 & \textbf{0.48} & 0.40 & & 0.47 & 0.45 & 0.47 & 0.42 & 0.46 & 0.46 \\
 & LPIPS$\downarrow$ & 0.43 & 0.34 & 0.37 & & 0.33 & 0.31 & 0.47 & 0.32 & 0.35 & \textbf{0.26} \\
\hline
\multirow{3}{*}{SCIE Part II} & PSNR$\uparrow$ & 19.90 & 19.60 & 19.92 & & 18.47 & 16.78 & 14.53 & 15.74 & \textbf{20.90} & 20.00 \\
 & SSIM$\uparrow$ & 0.73 & 0.77 & 0.77 & & 0.77 & 0.74 & 0.69 & 0.71 & \textbf{0.82} & 0.73 \\
 & LPIPS$\downarrow$ & 0.42 & 0.40 & 0.40 & & 0.40 & 0.40 & 0.53 & 0.39 & 0.32 & \textbf{0.16} \\
\hline
\end{tabular}
}
 \vspace{-3mm}
\label{main-r}
\end{table*}

\begin{table}[ht]
\centering
\caption{FLOPS and parameter counts for both supervised (SL) and unsupervised (UL) approaches.}
\vspace{0.1cm}
\begin{tabular}{clrr}
\hline
\textbf{Type} & \textbf{Method} & \textbf{FLOPS (G)} & \textbf{Params (M)} \\
\hline
\multirow{3}{*}{SL} & Retinex-Net~\cite{wei2018deep} & 587.470 & 0.555 \\
 & KinD~\cite{zhang2019kindling} & 574.954 & 8.160 \\
 & KinD++~\cite{zhang2021beyond} & 12238.030 & 8.275 \\
\hline
\multirow{6}{*}{UL} & EnlightenGan~\cite{jiang2021enlightengan} & 273.240 & 8.637 \\
 & ZeroDCE~\cite{li2021learning} & 84.990 & 0.079 \\
 & RUAS~\cite{liu2021retinex} & \textbf{3.528} & \textbf{0.003} \\
 & SCI~\cite{ma2022toward} & 188.873 & 0.011 \\
 & SRDACE~\cite{wen2023self} & 73.716 & 0.068 \\
 & \textbf{Ours} & 31.845 & 0.122 \\
\hline
\end{tabular}
\label{R-PF}
\end{table}



All methods are trained on SCIE Part I and evaluated in a zero-reference, cross-dataset setting to ensure a fair comparison between supervised and unsupervised approaches. Specifically, each model is trained on SCIE Part I and directly evaluated on LOLV1~\cite{wei2018deep}, LSRW~\cite{hai2023r2rnet}, and SCIE Part II~\cite{cai2018learning}, without any further fine-tuning or adaptation.
For SCIE Part II, we follow the evaluation protocol of~\cite{wen2023self} to construct the test split by selecting the first two images from each of the initial 125 folders and resizing them to $512 \times 512$ pixels.

\subsection{Implementation Details}


All models are trained using PyTorch with the AdamW optimizer on an NVIDIA RTX 3090 GPU, using a learning rate of $10^{-5}$, weight decay of $10^{-4}$, and a batch size of $8$. Model checkpoints are saved every 200 iterations for validation. We set $D = 7$, $S = 2$ and use the loss weights $\alpha : \beta : \gamma : \delta_1 : \delta_2 = 200 : 300 : 1 : 200 : 1000$. 

\subsection{Comparison Methods and Evaluation Metrics}

We benchmark our method against three supervised approaches (RetinexNet~\cite{wei2018deep}, KinD~\cite{zhang2019kindling}, and KinD++~\cite{zhang2021beyond}), which are evaluated under a zero-reference setting, as well as five unsupervised baselines (EnGAN~\cite{jiang2021enlightengan}, ZeroDCE~\cite{guo2020zero}, RUAS~\cite{liu2021retinex}, SCI~\cite{ma2022toward}, and SRDACE~\cite{wen2023self}). We evaluate performance using PSNR, SSIM~\cite{wang2004image}, and AlexNet-based LPIPS~\cite{zhang2018unreasonable}. In addition, we report the number of parameters and FLOPs. 

\subsection{Experimental Analysis}

\subsubsection{Qualitative Analysis}
Figure~\ref{fig:CMP} illustrates a challenging low-light scene in which the two ceramic vases and the supporting acrylic shelf are barely visible in the input image (a). While existing methods generally succeed in increasing overall brightness, they often introduce characteristic artifacts. For instance, KinD++ (b) enhances luminance but suffers from noticeable color casts and blocking artifacts. oURN (c) produces a cooler appearance, yet leaves important regions under-enhanced and introduces halo artifacts near sharp boundaries. SNR-Net (d) excessively amplifies contrast, leading to over-saturation and loss of fine texture. Retinexformer (e) and RRD (f) tend to oversmooth the scene or wash out subtle decorative patterns, resulting in a plastic-like appearance. ZeroDCE (g) preserves a relatively natural color tone but fails to sufficiently brighten shadowed regions. RUAS (i) and SCI (j) maintain smooth intensity transitions; however, they either suppress delicate details or retain residual darkness.


In contrast, our SIMI result (k) reconstructs the scene with: (1) faithful color rendition, preserving the original hues and saturation of the decorative motifs;
(2) artifact-free edge sharpness, yielding crisp contours without ringing effects; and
(3) balanced illumination, recovering natural mid-tones across both shadowed and highlighted areas. These improvements stem from the proposed bit-plane self-information mining strategy, which reveals structural and textural cues that are otherwise obscured by noise. By facilitating effective intrinsic and contextual feature fusion, SIMI alleviates the common artifact–performance trade-offs observed in the baseline methods (Fig.~\ref{fig:CMP}b–j), demonstrating strong qualitative robustness under the zero-reference setting.

\subsubsection{Quantitative Analysis}

Table~\ref{main-r} reports cross-dataset evaluation results on LOLV1, LSRW, and SCIE Part II, with all methods trained on SCIE Part I. SIMI achieves the highest PSNR on both LOLV1 and LSRW, outperforming all unsupervised competitors and surpassing the supervised baseline KinD++. Although SIMI does not obtain the highest SSIM on LOLV1, it consistently delivers superior perceptual quality, achieving the lowest LPIPS on LSRW and SCIE Part II. These results suggest that SIMI better preserves perceptually salient structures while maintaining strong pixel-level fidelity across datasets.

Table~\ref{R-PF} summarizes the computational cost and model size of the evaluated methods. SIMI requires 31.845 GFLOPS and 0.122 M parameters, offering a favorable balance between computational efficiency and representational capacity. Compared with RUAS (3.528 GFLOPS, 0.003 M parameters), SIMI incurs higher computational and parameter costs; however, this additional capacity translates into substantially improved restoration quality. At the same time, SIMI remains markedly more efficient than heavier supervised models such as KinD++ (12,238 GFLOPS), supporting its practical deployment.

\subsection{Ablation Studies}


\subsubsection{Usefulness of the self-information mining module}
We conduct an ablation study on LOLV1 to quantify the contribution of the proposed self-information mining module. As shown in Table~\ref{A-AW}, incorporating this module consistently enhances the base network, resulting in improvements of +0.51 dB in PSNR and +0.04 in SSIM. Beyond accuracy gains, the module also substantially accelerates optimization: the full model converges within 7,400 iterations, whereas removing the module leads to significantly slower convergence (Table~\ref{A-AW}). These results suggest that self-information mining provides more informative supervisory cues for learning enhancement transformations, thereby improving both reconstruction quality and training efficiency.




\subsubsection{Other decomposition methods}
To further validate the design choice of bit-plane decomposition, we compare it against two alternative decomposition strategies on LOLV1.

\textbf{Log-threshold decomposition.}
We define a set of logarithmically spaced thresholds $T=\{2^i\}_{i=0}^{7}$ and generate binary maps by thresholding each color channel as:
\begin{equation}
B_t^c=\mathbb{I}(I_0^c \ge t),\quad t\in T,
\end{equation}
where $\mathbb{I}(\cdot)$ denotes the indicator function.

\textbf{Quant-threshold decomposition.}
Alternatively, we uniformly quantize pixel intensities into $t_j$ discrete levels to obtain a coarsened representation:
\begin{equation}
I_{t_j}^c=\left\lfloor \frac{I_0^c(t_j-1)}{255}\right\rfloor \frac{255}{t_j-1}.
\end{equation}


As shown in Table~\ref{A-CDD}, the proposed bit-plane decomposition achieves the best overall performance, outperforming the log-threshold and quant-threshold variants by +0.92 dB and +0.61 dB in PSNR, respectively, while also attaining the highest SSIM. This advantage can be attributed to the ability of bit-plane decomposition to more effectively expose high-frequency structural cues, such as edges and fine textures, which are essential for perceptual fidelity in zero-reference enhancement scenarios. Overall, these results confirm that the proposed decomposition strategy plays a crucial role in enabling robust self-information mining and improving cross-dataset generalization.

\begin{table}[t!]
\centering
\caption{Ablation Study w/ and w/o Self-Information Mining Module (SIMM) on LOLV1 Dataset.}
\vspace{0.1cm}
\begin{tabular}{cccc}
\hline
\textbf{SIMI} & \textbf{PSNR$\uparrow$} & \textbf{SSIM$\uparrow$} & \textbf{Iteration$\downarrow$}\\
\hline
\checkmark & \textbf{18.8945} & \textbf{0.6016} & \textbf{7400} \\
$-$ & 18.3774  & 0.5602 & 423000 \\
\hline
\end{tabular}
\vspace{-3mm}
\label{A-AW}
\end{table}

\begin{table}[t!]
\centering
\caption{Comparison of Different Decomposition Approaches on LOLV1 Dataset.}
\vspace{0.1cm}
\begin{tabular}{ccccc}
\hline
\textbf{Metric} & \textbf{Ours} & \textbf{Log. thresh.} & \textbf{Qua. thresh.} \\
\hline
PSNR$\uparrow$ & \textbf{18.8945} & 17.9786 & 18.2782 \\
SSIM$\uparrow$ & \textbf{0.6016} & 0.5816 & 0.5687 \\
\hline
\end{tabular}
\label{A-CDD}
\end{table}

\section{Conclusion}


We propose \textbf{SIMI}, an unsupervised, zero-reference low-light enhancement method that mines self-information through bit-plane decomposition. This lightweight framework effectively extracts high-frequency structural cues (e.g., textures and boundaries) and subtle color-band variations that are otherwise obscured in low-light conditions. Experiments demonstrate state-of-the-art performance with minimal computational overhead

\bibliographystyle{IEEEbib}
\bibliography{refs}

\end{document}